\pdfoutput=1

\documentclass[11pt]{article}

\usepackage[]{ACL2023}

\usepackage{times}
\usepackage{latexsym}

\usepackage[T1]{fontenc}

\usepackage[utf8]{inputenc}

\usepackage{microtype}
\usepackage{latexsym}
\usepackage{microtype}
\usepackage{subfigure}
\usepackage{amssymb}
\usepackage{stfloats}
\usepackage{algorithm}
\usepackage{algorithmic}
\usepackage{float}
\usepackage{graphicx}
\usepackage{amsmath}
\usepackage{booktabs}
\usepackage{hyperref}

\usepackage{enumerate}
\usepackage{multirow}
\usepackage{epstopdf}

\usepackage{inconsolata}

\title{HiTIN: Hierarchy-aware Tree Isomorphism Network for \\ Hierarchical Text Classification}

\author{He Zhu$^{1*}$, Chong Zhang$^{1*}$,  Junjie Huang$^1$,  Junran Wu$^{1\dag}$, Ke Xu$^{1,2}$  \\
	$^1$State Key Lab of Software Development Environment\\
	Beihang University, Beijing, 100191, China\\
	$^{2}$Zhongguancun Laboratory, Beijing, 100094, China \\
	\{roy\_zh, chongzh, huangjunjie, wu\_junran, kexu\}@buaa.edu.cn}

\begin{document}
\maketitle
\begin{abstract}
Hierarchical text classification (HTC) is a challenging subtask of multi-label classification as the labels form a complex hierarchical structure. Existing dual-encoder methods in HTC achieve weak performance gains with huge memory overheads and their structure encoders heavily rely on domain knowledge. Under such observation, we tend to investigate the feasibility of a memory-friendly model with strong generalization capability that could boost the performance of HTC without prior statistics or label semantics. In this paper, we propose Hierarchy-aware Tree Isomorphism Network (HiTIN) to enhance the text representations with only syntactic information of the label hierarchy. Specifically, we convert the label hierarchy into an unweighted tree structure, termed coding tree, with the guidance of structural entropy. Then we design a structure encoder to incorporate hierarchy-aware information in the coding tree into text representations. Besides the text encoder, HiTIN only contains a few multi-layer perceptions and linear transformations, which greatly saves memory. We conduct experiments on three commonly used datasets and the results demonstrate that HiTIN could achieve better test performance and less memory consumption than state-of-the-art (SOTA) methods.

\renewcommand{\thefootnote}{}
\footnotetext{$^{*}$Equal Contribution.}
\footnotetext{$^{\dag}$Correspondence to: Junran Wu.}

\end{abstract}
\begin{figure}[!htb]
	\centering
	\includegraphics[width=0.48\textwidth]{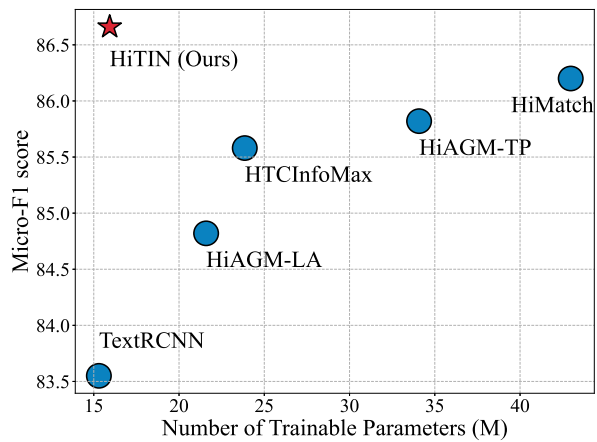}
	\caption{Micro-F1 score and the number of trainable parameters of our method and SOTAs with dual encoders on Web Of Science dataset.}
	\label{fig:tradeoff}
\end{figure}
\section{Introduction}
Hierarchical text classification is a sub-task of text multi-label classification, which is commonly applied in scenarios such as news document classification \citep{RCV1, NYT}, academic paper classification \citep{HDLTex}, and so on. Unlike traditional classification tasks, the labels of HTC have parent-child relationships forming a hierarchical structure. Due to the complex structure of label hierarchy and the imbalanced frequency of labels, HTC becomes a challenging task in natural language processing.

Recent studies in HTC typically utilize a dual-encoder framework \citep{HiAGM}, which consists of a text encoder for text representations and a structure encoder to inject the information of labels into text. The text encoder could be a traditional backbone for text classification, for instance, TextRCNN \citep{Lai2015TextRCNN} or BERT \citep{BERT}. The structure encoder is a Graph Neural Network (GNN) that treats the label hierarchy as a Directed Acyclic Graph (DAG) and propagates the information among labels. To maximize the propagation ability of the structure encoder, \citet{HiAGM} learn textual features of labels and count the prior probabilities between parent and child labels. Based on the dual-encoder framework, researchers further complicated the model by adding complementary networks and loss functions from different aspects, such as treating HTC as a matching problem \citep{HiMatch}, introducing mutual information maximization \citep{HTCInfoMax}. However, more complementary components result in more memory consumption, as shown in Figure~\ref{fig:tradeoff}. On the other hand, their structure encoders still rely on the prior statistics \citep{HiAGM, HiMatch} or the representation of labels \cite{HiAGM, HTCInfoMax}. That is, their models require a mass of domain knowledge, which greatly reduces the generalization ability.

To this end, we intend to design a more effective structure encoder with fewer parameters for HTC. Instead of introducing domain knowledge, we try to take full advantage of the structural information embedded in label hierarchies. Inspired by \citet{Li2016StructuralEntropy}, we decode the essential structure of label hierarchies into coding trees with the guidance of structural entropy, which aims to measure the structural complexity of a graph. The coding tree is unweighted and could reflect the hierarchical organization of the original graph, which provides us with another view of the label hierarchy. To construct coding trees, we design an algorithm, termed CodIng tRee Construction Algorithm (CIRCA) by minimizing the structural entropy of label hierarchies. Based on the hierarchical structure of coding trees, we propose Hierarchical-aware Tree Isomorphism Network (HiTIN). The document representations fetched by the text encoder are fed into a structure encoder, in which we iteratively update the node embeddings of the coding tree with a few multi-layer perceptions. Finally, we produce a feature vector of the entire coding tree as the final representation of the document. Compared with SOTA methods of dual encoders on HTC tasks \citep{HiAGM, HiMatch, HTCInfoMax, HGCLR}, HiTIN shows superior performance gains with less memory consumption. Overall, the contributions of our work can be summarized as follows:
\begin{itemize}
	\item To improve the generalization capability of dual-encoder models in HTC, we decode the essential structure of label hierarchies with the guidance of structural entropy. 
	\item We propose HiTIN, which has fewer learnable parameters and requires less domain knowledge, to fuse the structural information of label hierarchies into text representations.
	\item Numerous experiments are conducted on three benchmark datasets to demonstrate the superiority of our model. For reproducibility, our code is available at \href{https://github.com/Rooooyy/HiTIN}{https://github.com/Rooooyy/HiTIN}.
\end{itemize}
\section{Related Work}
\paragraph{Hierarchical Text Classification.}
Existing works for HTC could be categorized into local and global approaches \citep{HiAGM}. Local approaches build classifiers for a single label or labels at the same level in the hierarchy, while global approaches treat HTC as a flat classification task and build only one classifier for the entire taxonomy. Previous local studies mainly focus on transferring knowledge from models in the upper levels to models in the lower levels. \citet{HDLTex} first feed the whole corpus into the parent model and then input the documents with the same label marked by the parent model into a child model. In the next few years, researchers try different techniques to deliver knowledge from high-level models to low-level models \citep{Shimura2018HFTCNNLH,Huang2019HierarchicalMT,Banerjee2019HierarchicalTL}. 

Global studies in HTC try to improve flat multi-label classification by introducing various information from the hierarchy. \citet{Gopal2013RecursiveRF} propose a recursive regularization function to make the parameters of adjacent categories have similar values. \citet{Peng2018LargeScaleHT} propose a regularized graph-CNN model to capture the non-consecutive semantics from texts. Besides, various deep learning techniques, such as sequence-to-sequence model \citep{Yang2018SGMSG, Rojas2020}, attention mechanism \citep{You2019AttentionXML}, capsule network \citep{Aly2019HierarchicalMC, Peng2021HierarchicalTA}, reinforcement learning \citep{Mao2019HiRL}, and meta-learning \citep{Wu2019Meta} are also applied in global HTC. Recently, \citet{HiAGM} specially design an encoder for label hierarchies which could significantly improve performance. \citet{Chen2020HyperbolicIM} learn the word and label embeddings jointly in the hyperbolic space. \citet{HiMatch} formulate the text-label relationship as a semantic matching problem. \citet{HTCInfoMax} introduce information maximization which can model the interaction between text and label while filtering out irrelevant information. With the development of Pretrained Language Model (PLM), BERT\citep{BERT} based contrastive learning\citep{HGCLR}, prompt tuning\citep{HPT}, and other methods \citep{HBGL} have brought huge performance boost to HTC.

\paragraph{Structural Entropy.}
Structural entropy \citep{Li2016StructuralEntropy} is a natural extension of Shannon entropy \citep{Shannon} on graphs as structure entropy could measure the structural complexity of a graph. The structural entropy of a graph is defined as the average length of the codewords obtained by a random walk under a specific coding scheme. The coding scheme, termed coding tree \citep{Li2016StructuralEntropy}, is a tree structure that encodes and decodes the essential structure of the graph. In other words, to minimize structural entropy is to remove the noisy information from the graph. In the past few years, structural entropy has been successfully applied in network security \citep{Li2016ResistanceAS}, medicine \citep{Li2016ThreeDimensionalGM}, bioinformatics \citep{Li2018DecodingTA}, graph classification \citep{Wu2022HRN, wu2022structural}, text classification \citep{Zhang2022HINT}, and graph contrastive learning \citep{wu2023sega}.
\begin{figure*}[!th]
	\centering
	\includegraphics[width=\textwidth]{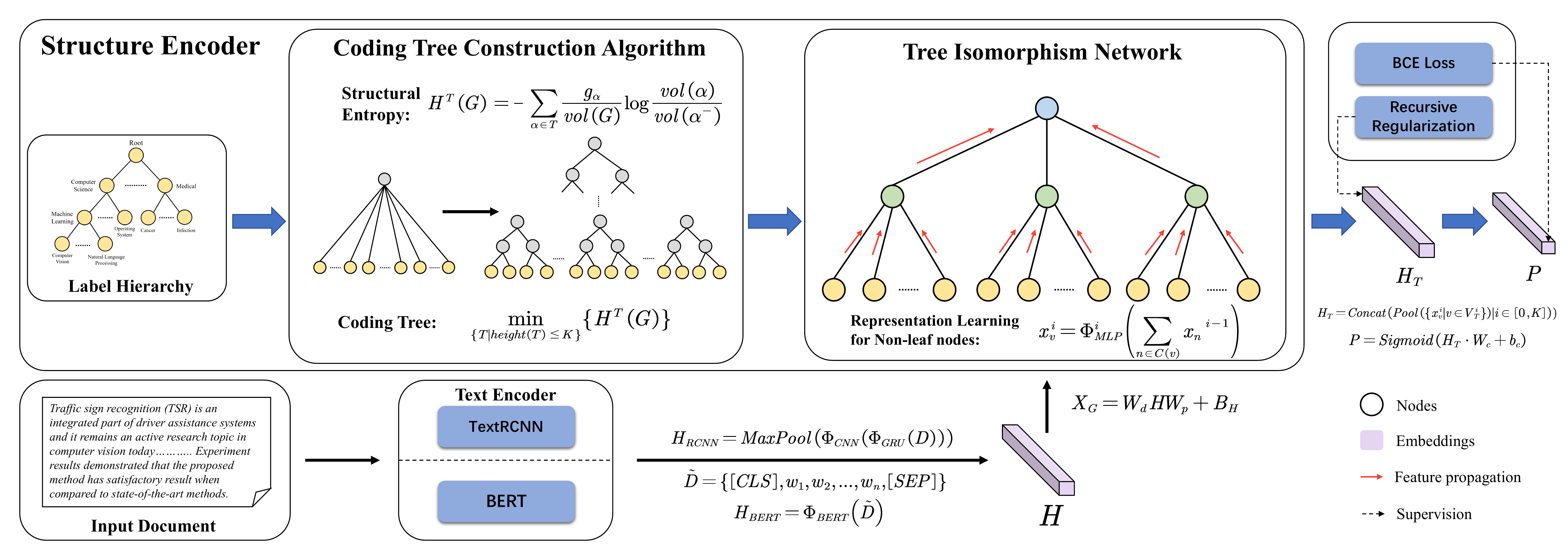}
	\caption{\label{fig:model}An example of HiTIN with $K=2$. As shown in Section~\ref{sec:text_encoder}, the input document is first fed into the text encoder to generate text representations. Next, the label hierarchy is transformed into a coding tree via Coding Tree Construction Algorithm proposed in Section~\ref{sec:structure_encoder}. The text representations are mapped into the leaf nodes of the coding tree and we iteratively update the non-leaf node embeddings in Section~\ref{sec:structure_encoder}. Finally, we produce a feature vector of the entire coding tree and calculate the classification probabilities in Section~\ref{sec:classification}. Besides, HiTIN is supervised by binary cross-entropy loss and recursive regularization \citep{Gopal2013RecursiveRF}. }
	\vspace{-2mm}
\end{figure*}
\section{Problem Definition}
Given a document $D=\{w_1, w_2, \dots, w_n\}$, where $w_i$ is a word and $n$ denotes the document length, hierarchical text classification aims to predict a subset $\mathcal{Y}$ of the holistic label set $Y$. Besides, every label in $Y$ corresponds to a unique node on a directed acyclic graph, i.e. the label hierarchy. The label hierarchy is predefined and usually simplified as a tree structure. In the groud-truth label set, a non-root label $y_i$ always co-occurs with its parent nodes, that is, for any $y_i \in \mathcal{Y}$, the parent node of $y_i$ is also in $\mathcal{Y}$.
\section{Methodology}
Following the dual-encoder scheme in HTC, the architecture of HiTIN that consists of a text encoder and a structure encoder is shown in Figure~\ref{fig:model}. The text encoder aims to capture textual information from the input document while the structure encoder could model the label correlations in the hierarchy and inject the information from labels into text representations.
\subsection{Text Encoder}
\label{sec:text_encoder}
In HTC, text encoder generally has two choices, that is, TextRCNN encoder and BERT encoder. TextRCNN \citep{Lai2015TextRCNN} is a traditional method in text classification, while BERT \citep{BERT} has shown its powerful ability in sequence feature extraction and has been widely applied in natural language processing in the past few years.
\paragraph{TextRCNN Encoder.}
The given document $D=\{w_1, w_2, \dots, w_n\}$, which is a sequence of word embeddings, is firstly fed into a bidirectional GRU layer to extract sequential information. Then, multiple CNN blocks along with max pooling over time are adopted to capture n-gram features. Formally,
\begin{equation}
	H_{RCNN} = MaxPool(\Phi_{CNN}(\Phi_{GRU}(D))),
\end{equation}
where $\Phi_{CNN}(\cdot)$ and $\Phi_{GRU}(\cdot)$ respectively denote a CNN and a GRU layer, while $MaxPool(\cdot)$ denotes the max pooling over time operation. Besides, $H_{RCNN} \in \mathbb{R} ^{n_C \times d_C},$ where $n_C$ denotes the number of CNN kernels and $d_C$ denotes the output channels of each CNN kernel.

The final representation $H\in \mathbb{R} ^ {n_C * d_C}$ of document $D$ is the concatenation of $H_{RCNN}$. That is,
\begin{equation}
	H = Concat(H_{RCNN}).
\end{equation}

\paragraph{BERT Encoder.}
Recent works in HTC also utilize BERT for learning textual features \citep{HiMatch, HGCLR}. Since there are few changes made to the vanilla BERT, we only introduce the workflow of our model and omit the details of BERT.

Given a input document $D=\{w_1, w_2, \dots, w_n\}$, we pad the document with two specical tokens:
\begin{equation}
	\tilde{D} = \{[CLS], w_1, w_2, \dots, w_n, [SEP]\},
\end{equation}
where $[CLS]$ and $[SEP]$ respectively denote the beginning and the end of the document. After padding and truncating, document $\tilde{D}$ is fed into BERT. Then BERT generates embeddings for each token in the document: 
\begin{equation}
	H_{BERT} = \Phi_{BERT}(\tilde{D}),
\end{equation}
where $H_{BERT} \in \mathbb{R}^{(n+2) \times d_B} $, and $\Phi_{BERT}(\cdot)$ denotes the BERT model. We adopt the CLS embedding as the representation of the entire text sequence. Thus, the final representation $H$ of document $D$ is:
\begin{equation}
	H = H_{BERT}^0, H \in \mathbb{R}^{d_B},
\end{equation}
where $d_B$ is the hidden dimension.
\subsection{Structure Encoder}
\label{sec:structure_encoder}
The semantic information provided by text encoder is then input into the structure encoder. Unlike previous works, we do not utilize the prior statistics or learn representations of the label hierarchy. Instead, we design a suite of methods guided by structural entropy \citep{Li2016StructuralEntropy} to effectively incorporate the information of text and labels.
\paragraph{Structural Entropy.}
 Inspired by \citet{Li2016StructuralEntropy}, we try to simplify the original structure of the label hierarchy by minimalizing its structural entropy. The structural entropy of a graph is defined as the average length of the codewords obtained by a random walk under a specific coding pattern named coding tree \citep{Li2016StructuralEntropy}. Given a graph $G=(V_G, E_G)$, the structural entropy of $G$ on coding tree $T$ is defined as:
\begin{equation}
	H^{T}(G)=-\sum_{\alpha \in T} \frac{g_{\alpha}}{vol(G)}\log{\frac{vol(\alpha)}{vol(\alpha^-)}},
\end{equation}
where $\alpha$ is a non-root node of coding tree $T$ which represents a subset of $V_G$, $\alpha^-$ is the parent node of $\alpha$ on the coding tree. $g_{\alpha}$ represents the number of edges with only one endpoint in $\alpha$ and the other end outside $\alpha$, that is, the out degree of $\alpha$. $vol(G)$ denotes the volume of graph $G$ while $vol(\alpha)$ and $vol(\alpha^-)$ is the sum of the degree of nodes that respectively partitioned by $\alpha$ and $\alpha^-$. 

For a certain coding pattern, the height of the coding tree should be fixed. Therefore, the $K$-dimensional structural entropy of the graph $G$ determined by the coding tree $T$ with a certain height $K$ is defined as:
\begin{equation}
	H_{K}(G)=\min_{\{T|height(T) \leq K\}}H^{T}(G).
\end{equation}
\paragraph{Coding Tree Construction Algorithm.}
To minimize the structural entropy of graph $G$, we design a CodIng tRee Construction Algorithm (CIRCA) to heuristically construct a coding tree $T$ with a certain height no greater than $K$. That is, $T = CIRCA(G, K)$, where $T=(V_T, E_T)$, $V_T = (V^0_T,\dots,V^h_T)$. To better illustrate CIRCA, we make some definitions as follows,
\newtheorem{myDef}{Definition}
\begin{myDef}
	\label{def:1}
	Let $T = (V_T, E_T)$ be a coding tree for graph $G=(V_G,E_G)$, $v_r$ be the root node of\quad$T$. For any $(v_i, v_j) \in T$, if $v_i$ is the direct child node of $v_j$, denote that
	\begin{equation*}
		v_i \in v_j.children;
	\end{equation*}
	and $v_j$ is equivalent to $v_i.parent$. 
\end{myDef} 
\begin{myDef}
	\label{def:2}
	Following Definition~\ref{def:1}, given any two nodes $(v_i, v_j) \in T$, in which $v_i \in v_r.children$ and $v_j \in v_r.children$.
	
	Define a member function $merge(v_i, v_j)$ of $T$. $T.merge(v_i, v_j)$ could insert a new node $v_{\epsilon}$ bewtween $v_r$ and $(v_i, v_j)$. Formally,
	\begin{align*}
		v_{\epsilon}.children &\leftarrow v_i;\\
		v_{\epsilon}.children &\leftarrow v_j;\\
		v_r.children &\leftarrow v_{\epsilon};\\
		V_T^{v_i.height + 1} \leftarrow v_{\epsilon};\quad&E_T \leftarrow (v_{\epsilon}, v_i),(v_{\epsilon}, v_j);
	\end{align*}
\end{myDef}
\begin{myDef}
	\label{def:3}
	Following Definition~\ref{def:1}, given a node $v_i$. Define a member function $delete(v_i)$ of $T$. $T.delete(v_i)$ could delete $v_i$ from $T$ and attach the child nodes of $v_i$ to its parent node. Formally,
	\begin{align*}
		v_i.&parent.children \leftarrow v_i.children;\\
		V_T &:= V_T - \{v_i\};\\
		E_T &:= E_T - \{(v_i.parent, v_i)\};\\
		E_T &:= E_T - \{(v_i, v) | v \in v_i.children\};\\
	\end{align*}
\end{myDef}
\begin{myDef}
	\label{def:4}
	Following Definition~\ref{def:1}, given any two nodes$(v_i, v_j)$, in which $v_i \in v_j.children$. Define a member function $shift(v_i, v_j)$ of $T$. $T.shift(v_i, v_j)$ could insert a new node $v_{\epsilon}$ between $v_i$ and $v_j$:
	\begin{align*}
		v_{\epsilon}.children \leftarrow v_i;&\quad v_j.children \leftarrow v_{\epsilon}; \\
		V_T^{v_i.height + 1} \leftarrow v_{\epsilon};&\quad E_T \leftarrow \{(v_j, v_{\epsilon}), (v_{\epsilon}, v_i)\};
	\end{align*}
\end{myDef}
Based on the above definitions, the pseudocode of CIRCA can be found in Algorithm~\ref{alg:CIRCA}. More details about coding trees and CIRCA are shown in Appendix~\ref{sec:appendix}.
\renewcommand{\thealgorithm}{\arabic{algorithm}}
\begin{algorithm}[!htb]
	\caption{Coding Tree Construction Algorithm}
	\label{alg:CIRCA}
	\textbf{Input}: A graph $G=(V_G,E_G)$ , a postive integer $K$\\
	\textbf{Output}: Coding tree $T=(V_T,E_T)$ of the graph $G$ with height $K$ 
	\begin{algorithmic}[1] 
		\STATE $V_T^0 := V$;\\
		\COMMENT {Stage 1: Construct a full-height binary-tree}
		\label{stg:1}
		\WHILE{$|v_r.children| > 2$}
		\label{stg:2}
		\STATE $(v_i, v_j) = argmax_{(v, v')}\{H^T(G) - H^{T.merge(v, v')}(G)\}$
		\label{stg:3}
		\STATE $T.merge(v_i, v_j)$
		\ENDWHILE
		
		\COMMENT {Stage 2: Squeeze $T$ to height $K$}
		\WHILE{$T.height > K$}
		\STATE $v_i = argmin_{v}\{H^{T.remove(v)}(G) - H^T(G)\}$
		\STATE $T.remove(v_i)$
		\ENDWHILE
		
		\COMMENT {Stage 3: Erase cross-layer links}
		\FOR{$v_i \in T$}
		\IF {$|v_i.parent.height - v_i.height| > 1$}
		\STATE $T.shift(v_i, v_i.parent)$
		\ENDIF
		\ENDFOR
		
		\STATE \textbf{return} $T$
	\end{algorithmic}
\end{algorithm}
\paragraph{Hierarchy-aware Tree Isomorphism Network.}
For representation learning, we reformulate the label hierarchy as a graph $G_L = (V_{G_L}, E_{G_L}, X_{G_L})$, where $V_{G_L}$, $E_{G_L}$ respectively denotes the node set and the edge set of $G_L$, $V_{G_L} = Y$ while $E_{G_L}$ is predefined in the corpus. In our work, $V_{G_L}$ and $E_{G_L}$ are represented by the unweighted adjacency matrix of $G_L$. $X_{G_L}$ is the node embedding matrix of $G_L$. Instead of learning the concept of labels, we directly broadcast the text representation to the label structure. Specifically, $X_G$ is transformed from the text representation $H$ by duplication and projection. Formally,
\begin{equation}
	X_G = W_d H W_p + B_H,
\end{equation}
where $W_d \in \mathbb{R} ^ {|Y| \times 1}$ and $W_p \in \mathbb{R} ^ {d_H * d_V}$ are learnable weights for the duplication and projection. $|Y|$ is the volume of the label set. $d_H$ and $d_V$ respectively denote the dimension of text and node. $B_H$ indicates the learnable bias and $B_H \in \mathbb{R}^{|Y| \times d_v}$.

Next, we simplify the structure of the label hierarchy into a coding tree with the guidance of structural entropy. Given a certain height $K$, the coding tree $T_L = (V_{T_L}, E_{T_L}, X_{T_L})$ of the label hierarchy could be constructed by CIRCA,
\begin{equation}
	(V_{T_L}, E_{T_L}) = CIRCA(G_L, K),
\end{equation}
where $V_{T_L} = \{V_{T_L}^0, V_{T_L}^1,...V_{T_L}^K\}$ are the layer-wise node sets of coding tree $T_L$ while  $X_{T_L} = \{X_{T_L}^0, X_{T_L}^1,...,X_{T_L}^K\}$ represents the node embeddings of $V_{T_L}^i$, $i \in [0, K]$.

The coding tree $T_L$ encodes and decodes the essential structure of $G_L$, which provides multi-granularity partitions for $G_L$. The root node $v_r$ is the roughest partition which represents the whole node set of $G_L$, so $V_{T_L}^K$ = $\{v_r\}$. For every node $v$ and its child nodes $\{v_1, v_2, \dots, v_z\}$, $v_1, v_2, \dots,$ and $v_z$ formulate a partition of $v$. Moreover, the leaf nodes in $T_L$ is an element-wise partition for $G_L$, that is, $V_{T_L}^0 = V_{G_L}$,  $X_{T_L}^0 = X_{G_L}$.

Note that $\{V_{T_L}^i | i \in [1, K]\}$ is given by CIRCA while their node embeddings $\{X_{T_L}^i | i \in [1, K]\}$ remain empty till now. Thus, we intend to update the un-fetched node representation of coding tree $T_L$. Following the message passing mechanism in Graph Isomorphism Network (GIN) \citep{GIN}, we design Hierarchy-aware Tree Isomorphism Network (HiTIN) according to the structure of coding trees.
For $x_{v}^i \in X_{T_L}^i$ in the $i$-th layer,
\begin{equation}
	x_v^i = \Phi_{MLP}^i(\sum\nolimits_{n\in C(v)} x_n^{i-1}),
\end{equation}
where $v \in V_T^i$, $x_v^i \in \mathbb{R} ^ {d_V}$ is the feature vector of node $v$, and $C(v)$ represents the child nodes of $v$ in coding tree $T_L$. $\Phi_{MLP}^i(\cdot)$ denotes a two-layer multi-layer perception within BatchNorm \citep{BatchNorm} and ReLU function. 
The learning stage starts from the leaf node (layer 0) and learns the representation of each node layer by layer until reaching the root node (layer $K$). 
Finally, a read-out function is applied to compute a representation of the entire coding tree $T_L$:
\begin{equation}
	\begin{split}
		H_T = Concat(Pool(\{x_v^i|v\in V_{T_L}^i\})\\| i \in [0, K])), 
	\end{split}
	\label{equ:treerep}
\end{equation}
where $Concat(\cdot)$ indicates the concatenation operation. $Pool(\cdot)$ in Eq.~\ref{equ:treerep} can be replaced with a summation, averaging, or maximization function. $H_T \in \mathbb{R} ^ {d_T}$ denotes the final representation of $T_L$.
\subsection{Classification and Loss Function}
\label{sec:classification}
Similar to previous studies \citep{HiAGM, HGCLR}, we flatten the hierarchy by attaching a unique multi-label classifier. $H_T$ is fed into a linear layer along with a sigmoid function to generate classification probability:
\begin{equation}
	P = Sigmoid(H_T \cdot W_c + b_c),
\end{equation}
where $W_c \in \mathbb{R} ^ {d_T \times |Y|}$ and $b_c \in \mathbb{R} ^ {|Y|}$ are weights and bias of linear layer while $|Y|$ is the volume of the label set. For multi-label classification, we adopt the Binary Cross-Entropy Loss as the classification loss:
\begin{equation}
	\small{L^{C} = - \frac{1}{|Y|} \sum_{j}^{|Y|} y_{j}log(p_{j}) + (1 - y_{j})log(1 - p_{j}),}
\end{equation}
where $y_j$ is the ground truth of the $j$-th label while $p_j$ is the $j$-th element of $P$. Considering hierarchical classification, we use recursive regularization \citet{Gopal2013RecursiveRF} to constrain the weights of adjacent classes to be in the same distributions as formulated in Eq.~\ref{eq:rr}:
\begin{equation}
	L^{R} = \sum_{p \in Y} \sum_{q \in child(p)} \frac{1}{2} ||w_p^{2} - w_q^{2}||,
\label{eq:rr}
\end{equation}
where $p$ is a non-leaf label in $Y$ and $q$ is a child of $p$. $w_p, w_q \in W_c$. We use a hyper-parameter $\lambda$ to control the strength of recursive regularization. Thus, the final loss function can be formulated as:
\begin{equation}
	L = L^{C} + \lambda \cdot L^{R}.
\end{equation}
\vspace{-10pt}
\begin{table}[!ht]
	\vspace{-15pt}
	\centering
	\resizebox{0.48\textwidth}{!}{
		\begin{tabular}{ccccccc}
			\hline
			Dataset & 
			$|Y|$ &
			$Avg(y_i)$ & 
			Depth & 
			\# Train & 
			\# Dev & 
			\# Test \\ \hline
			WOS & 
			141	&	2.0	&	2 &	30,070 &	7,518 &	9,397            \\
			RCV1-v2 & 
			103	&	3.24 &	4 & 20,833 & 	2,316 & 781,265          \\
			NYTimes & 
			166 &	7.6 &	8 & 23,345 &	5,834 & 7,292            \\ \hline \hline
	\end{tabular}}
	\caption{Summary statistics of datasets.}
	\label{tab:data_stat}
\end{table}
\begin{table*}[!th]
	\centering
	\resizebox{\textwidth}{!}{
		\begin{tabular}{lcccccccc}
			\toprule[1pt]
			\multicolumn{1}{c}{\multirow{2}{*}{Hierarchy-aware Models}} & \multicolumn{2}{c}{WOS}      & 
			\multicolumn{2}{c}{RCV1-v2}        & 
			\multicolumn{2}{c}{NYTimes}         &
			\multicolumn{2}{c}{Average}        \\ \cline{2-9} 
			\multicolumn{1}{c}{}                       
			& Micro-F1       & Macro-F1             
			& Micro-F1       & Macro-F1             
			& Micro-F1       & Macro-F1             
			& Micro-F1       & Macro-F1
			\\ \hline                                                  
			TextRCNN \citep{HiAGM}                                   
			& 83.55          & 76.99
			& 81.57          & 59.25                
			& 70.83          & 56.18                
			& 78.65			 & 64.14
			\\
			HiAGM \citep{HiAGM}                                      
			& 85.82          & 80.28                
			& 83.96          & 63.35                
			& 74.97          & 60.83                
			& 81.58			 & 68.15
			\\
			HTCInfoMax \citep{HTCInfoMax}                                 
			& 85.58          & 80.05                
			& 83.51          & 62.71                
			& -              & -                    
			& -		  	     & -
			\\
			HiMatch \citep{HiMatch}                                    
			& 86.20          & 80.53                
			& 84.73       	 & 64.11                
			& -              & -                    
			& - 		     & -
			\\ \hline
			HiTIN                                       
			& \textbf{86.66} & \textbf{81.11}  
			& \textbf{84.81} & \textbf{64.37}       
			& \textbf{75.13} & \textbf{61.09}
			& \textbf{82.20} &\textbf{68.86}       
			\\ \bottomrule[1pt]
	\end{tabular}}
	\caption{Main Experimental Results with TextRCNN encoders. All baselines above and our method utilize GloVe embeddings \cite{glove} to initialize documents and encode them with TextRCNN \citep{Lai2015TextRCNN}.}
	\label{tab:main_glove}
\end{table*}
\begin{table*}[!th]
	\centering
	\resizebox{\textwidth}{!}{
		\begin{tabular}{lcccccccc}
			\toprule[1pt]
			\multicolumn{1}{c}{\multirow{2}{*}{Pretrained Language Models}} & \multicolumn{2}{c}{WOS}      & 
			\multicolumn{2}{c}{RCV1-v2}        & 
			\multicolumn{2}{c}{NYTimes}        & 
			\multicolumn{2}{c}{Average}       \\ \cline{2-9} 
			\multicolumn{1}{c}{}                       
			& Micro-F1       & Macro-F1             
			& Micro-F1       & Macro-F1             
			& Micro-F1       & Macro-F1
			& Micro-F1       & Macro-F1             
			\\ \hline		
			BERT $\dagger$                               
			& 85.63          & 79.07                
			& 85.65          & 67.02                
			& 78.24          & 65.62                
			& 83.17			 & 70.57
			\\
			BERT+HiAGM$\dagger$                             
			& 86.04          & 80.19                
			& 85.58          & 67.93                
			& 78.64          & 66.76                
			& 83.42			 & 71.63
			\\
			BERT+HTCInfoMax$\dagger$                            
			& 86.30          & 79.97                
			& 85.53          & 67.09                
			& 78.75          & 67.31                
			& 83.53 		 & 71.46 
			\\
			BERT+HiMatch \citep{HiMatch}                               
			& 86.70          & 81.06                
			& 86.33          & 68.66                
			& -              & -                    
			& -				 & -
			\\
			HGCLR \citep{HGCLR}                                      
			& 87.11 		 & 81.20       
			& 86.49          & 68.31                
			& 78.86          & 67.96                
			& 84.15			 & 72.49
			\\ \hline
			HiTIN                                       
			& \textbf{87.19} & \textbf{81.57}                
			& \textbf{86.71} & \textbf{69.95}       
			& \textbf{79.65} & \textbf{69.31}       
			& \textbf{84.52} & \textbf{73.61}
			\\ \bottomrule[1pt]
	\end{tabular}}
	\caption{Main Experimental Results with BERT encoder. All baselines above and our method adopt BERT\cite{BERT} as the text encoder. $\dagger$ denotes the results are reported by \citet{HGCLR}.}
	\label{tab:main_bert}
\end{table*}
\section{Experiments}
\subsection{Experiment Setup}
\paragraph{Datasets and Evaluation Metrics.}
We conduct experiments on three benchmark datasets in HTC. RCV1-v2 \citep{RCV1} and NYT \citep{NYT} respectively consist of news articles published by Reuters, Ltd. and New York Times, while WOS \citep{HDLTex} includes abstracts of academic papers from Web of Science. Each of these datasets is annotated with ground-truth labels in a given hierarchy. We split and preprocess these datasets following \citet{HiAGM}. The statistics of these datasets are shown in Table~\ref{tab:data_stat}. The experimental results are measured with Micro-F1 and Macro-F1 \citep{Gopal2013RecursiveRF}. Micro-F1 is the harmonic mean of the overall precision and recall of all the test instances, while Macro-F1 is the average F1-score of each category. Thus, Micro-F1 reflects the performance on more frequent labels, while Macro-F1 treats labels equally.
\paragraph{Implementation Details.}
The text embeddings fed into the TextRCNN encoder are initialized with GloVe \citep{glove}. The TextRCNN encoder consists of a two-layer BiGRU with hidden dimension 128 and CNN layers with kernel size=[2, 3, 4] and $d_C$=100. Thus, the hidden dimension of the final text representation is $d_H = r_C * d_C = 3 * 100 = 300$. The height $K$ of the coding tree is 2 for all three datasets. The hidden dimension $d_V$ of node embedding $X_G$ is set to 512 for RCV1-v2 while 300 for WOS and NYTimes. $Pool(\cdot)$ in Eq.~\ref{equ:treerep} is summation for all the datasets. The balance factor $\lambda$ for $L^R$ is set to 1e-6. The batch size is set to 16 for RCV1-v2 and 64 for WOS and NYTimes. The model is optimized by Adam \citep{Adam} with a learning rate of 1e-4.

For BERT text encoder, we use the BertModel of \texttt{bert-base-uncased} and there are some negligible changes to make it compatible with our method. $d_B = d_H = d_V = 768$. The height $K$ of the coding tree is 2 and the $Pool(\cdot)$ in Eq.~\ref{equ:treerep} is averaging. The batch size is set to 12, and the BertModel is fine-tuned by Adam \citep{Adam} with a learning rate of 2e-5.

\paragraph{Baselines.}
We compare HiTIN with SOTAs including HiAGM\citep{HiAGM}, HTCInfoMax \citep{HTCInfoMax}, HiMatch \citep{HiMatch}, and HGCLR \citep{HGCLR}. HiAGM, HTCInfoMax, and HiMatch use different fusion strategies to model text-hierarchy correlations. Specifically, HiAGM proposes a multi-label attention and a text feature propagation technique to get hierarchy-aware representations. HTCInfoMax enhances HiAGM-LA with information maximization to model the interaction between text and hierarchy. HiMatch treats HTC as a matching problem by mapping text and labels into a joint embedding space. HGCLR directly incorporates hierarchy into BERT with contrastive learning. 

\subsection{Experimental Results}
The experimental results with different types of text encoders are shown in Table~\ref{tab:main_glove} and Table~\ref{tab:main_bert}. HiAGM is the first method to apply the dual-encoder framework and outperforms TextRCNN on all the datasets. HTCInfoMax improves HiAGM-LA \citep{HiAGM} by introducing mutual information maximization but is still weaker than HiAGM-TP. HiMatch treats HTC as a matching problem and surpasses HiAGM-TP\citep{HiAGM} on WOS and RCV1-v2. Different from these methods, HiTIN could further extract the information in the text without counting the prior probabilities between parent and child labels or building feature vectors for labels. As shown in Table~\ref{tab:main_glove}, when using TextRCNN as the text encoder, our model outperforms all baselines on the three datasets. Based on TextRCNN, HiTIN brings 3.55\% and 4.72\% improvement of Micro-F1 and Macro-F1 on average.

As for pretrained models in Table~\ref{tab:main_bert}, our model also beats existing methods in all three datasets. Compared with vanilla BERT, our model can significantly refine the text representations by respectively achieving 1.2\% and 3.1\% average improvement of Micro-F1 and Macro-F1 on the three datasets. In addition, our method can achieve 3.69\% improvement of Macro-F1 on NYT, which has the deepest label hierarchy in the three datasets. It demonstrates the superiority of our model on the dataset with a complex hierarchy. Compared with BERT-based HTC methods, our model observes a 1.12\% average improvement of Macro-F1 against HGCLR. On RCV1-v2, the performance boost of Macro-F1 even reaches 1.64\%. The improvement of Macro-F1 shows that our model could effectively capture the correlation between parent and child labels even without their prior probabilities.
\begin{table}[!th]
	\centering
	\resizebox{0.48\textwidth}{!}{
		\begin{tabular}{ccccccc}
			\hline
			\multirow{2}{*}{Ablation Models} 
			& \multicolumn{2}{c}{WOS}         
			& \multicolumn{2}{c}{RCV1-v2}
			& \multicolumn{2}{c}{NYTimes}        
			\\ \cline{2-7} 
			& Micro-F1       & Macro-F1       
			& Micro-F1       & Macro-F1       
			& Micro-F1       & Macro-F1 \\ \hline
			HiTIN(Random)          
			& 84.74          & 77.90          
			& 82.41          & 61.46          
			&71.99			 & 58.26 \\ 
			w/o $L^R$                 
			& 86.48          & 80.48          
			& 84.14          & 63.12          
			& 74.93			 & 59.95 \\
			HiTIN                  
			& \textbf{86.66} & \textbf{81.11} 
			& \textbf{84.81} & \textbf{64.37} 
			& \textbf{75.13} &\textbf{61.09} \\ \hline \hline
	\end{tabular}}
	\caption{Performance when replacing or removing a component of HiTIN. HiTIN(Random) denotes the results produced by HiTIN within the random algorithm. w/o $L^R$ stands for the parameter $\lambda$ is set to 0.}
	\label{tab:ablation}
	\vspace{-10pt}
\end{table}
\subsection{The Necessity of CIRCA}
In this subsection, we illustrate the effectiveness of CIRCA by comparing it to a random algorithm. The random algorithm generates a coding tree of the original graph $G$ with a certain height $K$ just like CIRCA. First, the random algorithm also takes all nodes of graph $G$ as leaf nodes of the tree. But different from CIRCA, for each layer, every two nodes are randomly paired and then connect to their parent node. Finally, all nodes in the ${K-1}_{th}$ layer are connected to a root node. We generate coding trees with the random algorithm and then feed them into our model. 

As shown in Table~\ref{tab:ablation}, the results demonstrate that the random algorithm leads to a negative impact which destroys the original semantic information. Thus, it is difficult for the downstream model to extract useful features. On the contrary, the coding tree constructed by CIRCA can retain the essential structure of the label hierarchy and make the learning procedure more effective. Besides, our model could achieve good performance without Eq.~\ref{eq:rr}, which proves that CIRCA could retain the information of low-frequency labels while minimizing the structural entropy of label hierarchies.
\begin{figure*}[!htp]
	\centering
	\subfigure[WOS]{\includegraphics[width=0.3\textwidth]{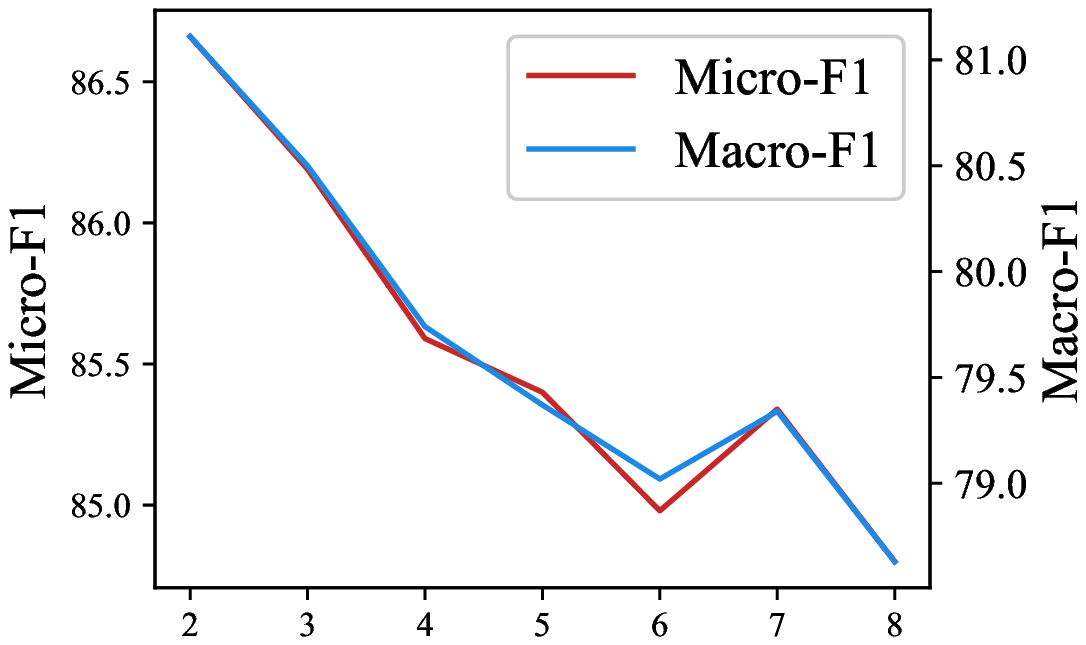}}
	\subfigure[RCV1-v2]{\includegraphics[width=0.3\textwidth]{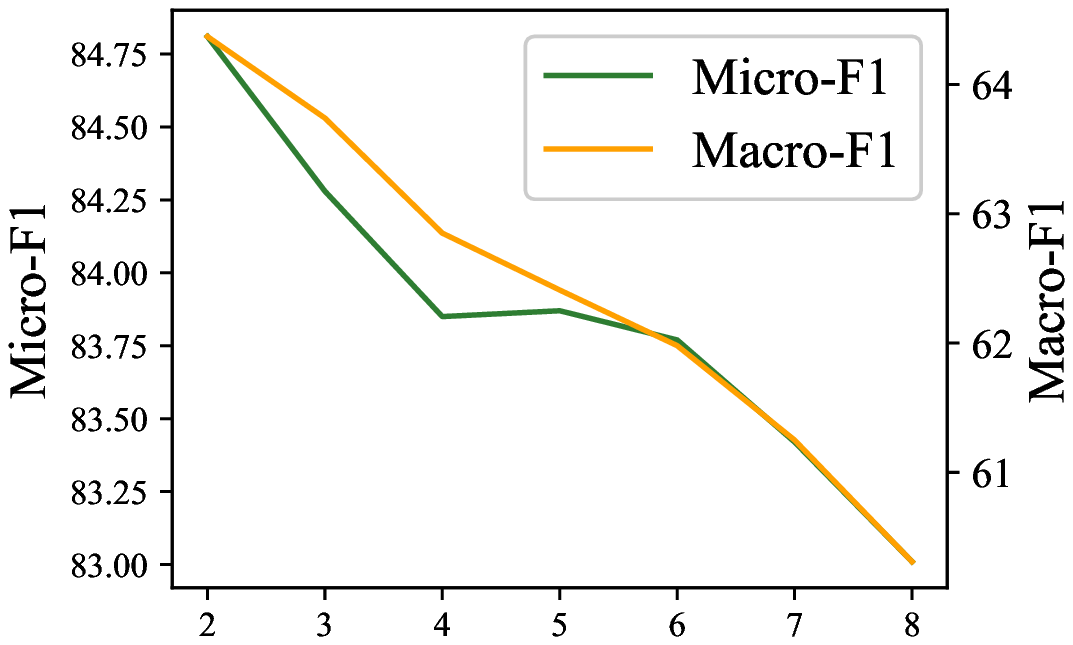}}
	\subfigure[NYTimes]{\includegraphics[width=0.3\textwidth]{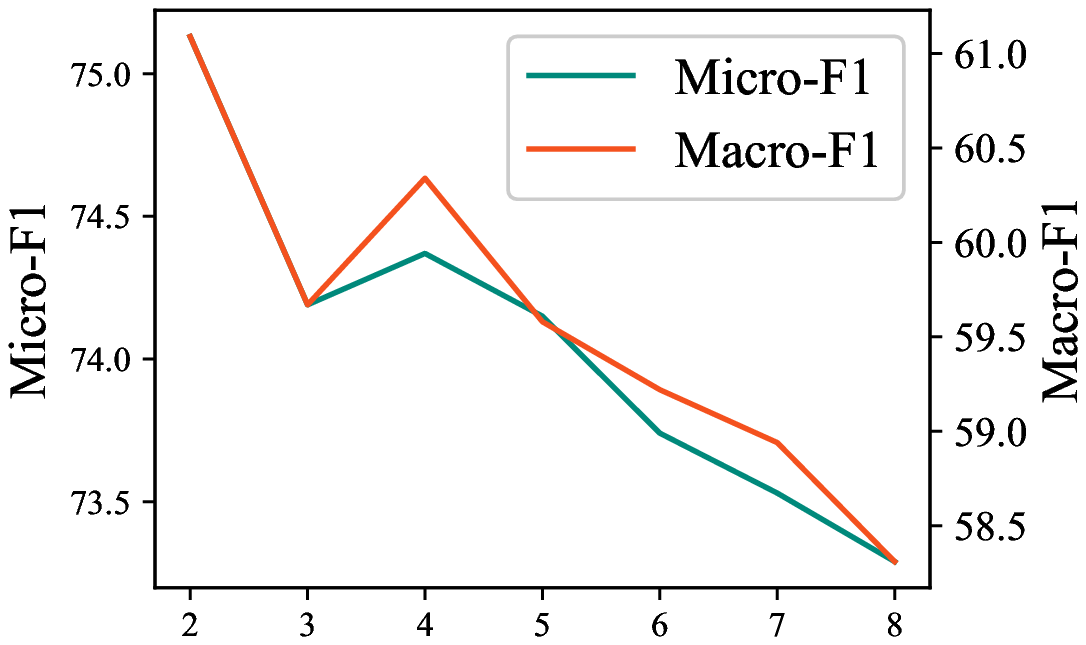}}
	\caption{Test performance of HiTIN with different height $K$ of the coding tree on three datasets.}
	\label{fig:k}
	\vspace{-2mm}
\end{figure*}
\subsection{The Height of Coding Tree}
The height of the coding tree directly affects the performance of our model. The higher the coding tree, the more information is compressed. To investigate the impact of $K$, we run HiTIN with different heights $K$ of the coding tree while keeping other settings the same. Figure~\ref{fig:k} shows the test performance of different height coding trees on WOS, RCV1-v2, and NYTimes. As $K$ grows, the performance of HiTIN is severely degraded. Despite the different depths of label hierarchy, the optimal heights of the coding tree for the three datasets are always 2. A probable reason is that the 2-dimensional structural entropy roughly corresponds to objects in the 2-dimensional space as the text and label are both represented with 2-D tensors. On the other hand, as $K$ grows, more noisy information is eliminated, but more useful information is also compressed.
\begin{figure}[!htb]
	\centering
	\includegraphics[width=0.86\columnwidth]{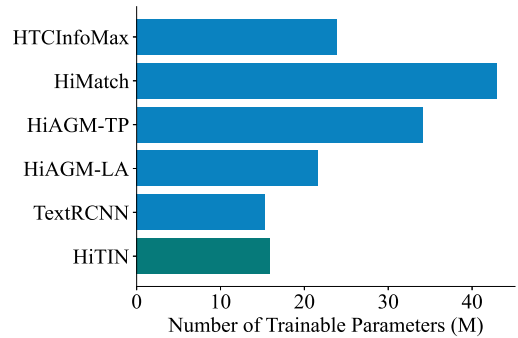}
	\caption{The number of trainable parameters of HiTIN and baseline models on WOS.}
	\label{fig:param}
	\vspace{-10pt}
\end{figure}
\subsection{The Mermory-saving Feature of HiTIN}
In this subsection, we compare the number of learnable parameters of HiTIN with that of the baselines. We set $K$ to 2 and run these models on WOS while keeping the other hyper-parameter the same. The numbers of trainable parameters are counted by the $numel(\cdot)$ function in PyTorch \citep{Pytorch}. As shown in Figure~\ref{fig:param}, we can observe that the parameter of our model is slightly greater than TextRCNN \citep{HiAGM} but significantly smaller than HiAGM \citep{HiAGM}, HiMatch \citep{HiMatch}, and HTCInfoMax \citep{HTCInfoMax}. One important reason is the simple and efficient architecture of HiTIN, which contains only a few MLPs and linear transformations. On the contrary, HiAGM-LA \citep{HiAGM} needs extra memory for label representations, HiAGM-TP uses a space-consuming method for text-to-label transformation, and both of them utilized gated network as the structure encoder, which further aggravates memory usage. HiMatch \citep{HiMatch} and HTCInforMax \citep{HTCInfoMax} respectively introduce auxiliary neural networks based on HiAGM-TP and HiAGM-LA. Thus, their memory usages are even larger.

\section{Conclusion}
In this paper, we propose a suite of methods to address the limitations of existing approaches regarding HTC. In particular, tending to minimize structural entropy, we design CIRCA to construct coding trees for the label hierarchy. To further extract textual information, we propose HiTIN to update node embeddings of the coding tree iteratively. Experimental results demonstrate that HiTIN could enhance text representations with only structural information of the label hierarchy. Our model outperforms existing methods while greatly reducing memory increments.

\section*{Limitations}
For text classification tasks, the text encoder is more important than other components. Due to the lack of label semantic information and simplified learning procedure, the robustness of text encoders directly affects the performance of our model. From Table~\ref{tab:main_glove} and~\ref{tab:main_bert}, we could observe that BERT has already surpassed TextRCNN by 4.52\% and 6.43\% on Micro-F1 and Macro-F1. Besides, BERT beats all the TextRCNN-based methods on RCV1-v2 and NYTimes. However, when applying BERT as the text encoder, our model makes slight improvements to Micro-F1, especially on WOS. A probable reason is that BERT was pre-trained on news corpus while WOS consists of academic papers.


\section*{Acknowledgements}
This research was supported by NSFC (Grant No. 61932002).

\bibliography{HTC}

\begin{thebibliography}{38}
\expandafter\ifx\csname natexlab\endcsname\relax\def\natexlab#1{#1}\fi

\bibitem[{Aly et~al.(2019)Aly, Remus, and Biemann}]{Aly2019HierarchicalMC}
Rami Aly, Steffen Remus, and Chris Biemann. 2019.
\newblock Hierarchical multi-label classification of text with capsule
  networks.
\newblock In \emph{ACL}.

\bibitem[{Banerjee et~al.(2019)Banerjee, Akkaya, Perez-Sorrosal, and
  Tsioutsiouliklis}]{Banerjee2019HierarchicalTL}
Siddhartha Banerjee, Cem Akkaya, Francisco Perez-Sorrosal, and Kostas
  Tsioutsiouliklis. 2019.
\newblock Hierarchical transfer learning for multi-label text classification.
\newblock In \emph{ACL}.

\bibitem[{Chen et~al.(2020)Chen, Huang, Xiao, Cai, and
  Jing}]{Chen2020HyperbolicIM}
Boli Chen, Xin Huang, Lin Xiao, Zixin Cai, and Liping Jing. 2020.
\newblock Hyperbolic interaction model for hierarchical multi-label
  classification.
\newblock In \emph{AAAI}.

\bibitem[{Chen et~al.(2021)Chen, Ma, Lin, and Yan}]{HiMatch}
Haibin Chen, Qianli Ma, Zhenxi Lin, and Jiangyue Yan. 2021.
\newblock Hierarchy-aware label semantics matching network for hierarchical
  text classification.
\newblock In \emph{ACL}.

\bibitem[{Deng et~al.(2021)Deng, Peng, He, Li, and Yu}]{HTCInfoMax}
Zhongfen Deng, Hao Peng, Dongxiao He, Jianxin Li, and Philip~S. Yu. 2021.
\newblock \href {https://doi.org/10.18653/v1/2021.naacl-main.260} {Htcinfomax:
  {A} global model for hierarchical text classification via information
  maximization}.
\newblock In \emph{Proceedings of the 2021 Conference of the North American
  Chapter of the Association for Computational Linguistics: Human Language
  Technologies, {NAACL-HLT} 2021, Online, June 6-11, 2021}, pages 3259--3265.
  Association for Computational Linguistics.

\bibitem[{Devlin et~al.(2019)Devlin, Chang, Lee, and Toutanova}]{BERT}
Jacob Devlin, Ming{-}Wei Chang, Kenton Lee, and Kristina Toutanova. 2019.
\newblock \href {https://doi.org/10.18653/v1/n19-1423} {{BERT:} pre-training of
  deep bidirectional transformers for language understanding}.
\newblock In \emph{Proceedings of the 2019 Conference of the North American
  Chapter of the Association for Computational Linguistics: Human Language
  Technologies, {NAACL-HLT} 2019, Minneapolis, MN, USA, June 2-7, 2019, Volume
  1 (Long and Short Papers)}, pages 4171--4186. Association for Computational
  Linguistics.

\bibitem[{Gopal and Yang(2013)}]{Gopal2013RecursiveRF}
Siddharth Gopal and Yiming Yang. 2013.
\newblock Recursive regularization for large-scale classification with
  hierarchical and graphical dependencies.
\newblock \emph{Proceedings of the 19th ACM SIGKDD international conference on
  Knowledge discovery and data mining}.

\bibitem[{Huang et~al.(2019)Huang, Chen, Liu, Chen, Huang, Liu, Zhao, Zhang,
  and Wang}]{Huang2019HierarchicalMT}
Wei Huang, Enhong Chen, Qi~Liu, Yuying Chen, Zai Huang, Yang Liu, Zhou Zhao,
  Dandan Zhang, and Shijin Wang. 2019.
\newblock Hierarchical multi-label text classification: An attention-based
  recurrent network approach.
\newblock \emph{Proceedings of the 28th ACM International Conference on
  Information and Knowledge Management}.

\bibitem[{Ioffe and Szegedy(2015)}]{BatchNorm}
Sergey Ioffe and Christian Szegedy. 2015.
\newblock \href {http://proceedings.mlr.press/v37/ioffe15.html} {Batch
  normalization: Accelerating deep network training by reducing internal
  covariate shift}.
\newblock In \emph{Proceedings of the 32nd International Conference on Machine
  Learning, {ICML} 2015, Lille, France, 6-11 July 2015}, volume~37 of
  \emph{{JMLR} Workshop and Conference Proceedings}, pages 448--456. JMLR.org.

\bibitem[{Jiang et~al.(2022)Jiang, Wang, Sun, Chen, Zhuang, and Yang}]{HBGL}
Ting Jiang, Deqing Wang, Leilei Sun, Zhong-Yong Chen, Fuzhen Zhuang, and
  Qinghong Yang. 2022.
\newblock Exploiting global and local hierarchies for hierarchical text
  classification.

\bibitem[{Kingma and Ba(2014)}]{Adam}
Diederik~P. Kingma and Jimmy Ba. 2014.
\newblock Adam: A method for stochastic optimization.
\newblock \emph{CoRR}, abs/1412.6980.

\bibitem[{Kowsari et~al.(2017)Kowsari, Brown, Heidarysafa, Meimandi, Gerber,
  and Barnes}]{HDLTex}
Kamran Kowsari, Donald~E. Brown, Mojtaba Heidarysafa, K.~Meimandi, Matthew~S.
  Gerber, and Laura~E. Barnes. 2017.
\newblock Hdltex: Hierarchical deep learning for text classification.
\newblock \emph{2017 16th IEEE International Conference on Machine Learning and
  Applications (ICMLA)}, pages 364--371.

\bibitem[{Lai et~al.(2015)Lai, Xu, Liu, and Zhao}]{Lai2015TextRCNN}
Siwei Lai, Liheng Xu, Kang Liu, and Jun Zhao. 2015.
\newblock Recurrent convolutional neural networks for text classification.
\newblock In \emph{AAAI}.

\bibitem[{Lewis et~al.(2004)Lewis, Yang, Rose, and Li}]{RCV1}
David~D. Lewis, Yiming Yang, Tony~G. Rose, and Fan Li. 2004.
\newblock Rcv1: A new benchmark collection for text categorization research.
\newblock \emph{J. Mach. Learn. Res.}, 5:361--397.

\bibitem[{Li et~al.(2016{\natexlab{a}})Li, Hu, Liu, and
  Pan}]{Li2016ResistanceAS}
Angsheng Li, Qifu Hu, Jun Liu, and Yicheng Pan. 2016{\natexlab{a}}.
\newblock Resistance and security index of networks: Structural information
  perspective of network security.
\newblock \emph{Scientific Reports}, 6.

\bibitem[{Li and Pan(2016)}]{Li2016StructuralEntropy}
Angsheng Li and Yicheng Pan. 2016.
\newblock Structural information and dynamical complexity of networks.
\newblock \emph{IEEE Transactions on Information Theory}, 62:3290--3339.

\bibitem[{Li et~al.(2016{\natexlab{b}})Li, Yin, and
  Pan}]{Li2016ThreeDimensionalGM}
Angsheng Li, Xianchen Yin, and Yicheng Pan. 2016{\natexlab{b}}.
\newblock Three-dimensional gene map of cancer cell types: Structural entropy
  minimisation principle for defining tumour subtypes.
\newblock \emph{Scientific Reports}, 6.

\bibitem[{Li et~al.(2018)Li, Yin, Xu, Wang, Han, Wei, Deng, Xiong, and
  Zhang}]{Li2018DecodingTA}
Angsheng Li, Xianchen Yin, Bingxian Xu, Danyang Wang, Jimin Han, Yi~Wei, Yun
  Deng, Yingluo Xiong, and Zhihua Zhang. 2018.
\newblock Decoding topologically associating domains with ultra-low resolution
  hi-c data by graph structural entropy.
\newblock \emph{Nature Communications}, 9.

\bibitem[{Mao et~al.(2019)Mao, Tian, Han, and Ren}]{Mao2019HiRL}
Yuning Mao, Jingjing Tian, Jiawei Han, and Xiang Ren. 2019.
\newblock \href {https://doi.org/10.18653/v1/D19-1042} {Hierarchical text
  classification with reinforced label assignment}.
\newblock In \emph{Proceedings of the 2019 Conference on Empirical Methods in
  Natural Language Processing and the 9th International Joint Conference on
  Natural Language Processing, {EMNLP-IJCNLP} 2019, Hong Kong, China, November
  3-7, 2019}, pages 445--455. Association for Computational Linguistics.

\bibitem[{Paszke et~al.(2019)Paszke, Gross, Massa, Lerer, Bradbury, Chanan,
  Killeen, Lin, Gimelshein, Antiga, Desmaison, Kopf, Yang, DeVito, Raison,
  Tejani, Chilamkurthy, Steiner, Fang, Bai, and Chintala}]{Pytorch}
Adam Paszke, Sam Gross, Francisco Massa, Adam Lerer, James Bradbury, Gregory
  Chanan, Trevor Killeen, Zeming Lin, Natalia Gimelshein, Luca Antiga, Alban
  Desmaison, Andreas Kopf, Edward Yang, Zachary DeVito, Martin Raison, Alykhan
  Tejani, Sasank Chilamkurthy, Benoit Steiner, Lu~Fang, Junjie Bai, and Soumith
  Chintala. 2019.
\newblock \href
  {http://papers.neurips.cc/paper/9015-pytorch-an-imperative-style-high-performance-deep-learning-library.pdf}
  {{PyTorch: An Imperative Style, High-Performance Deep Learning Library}}.
\newblock In \emph{Advances in Neural Information Processing Systems 32}, pages
  8024--8035. Curran Associates, Inc.

\bibitem[{Peng et~al.(2021)Peng, Li, Gong, Wang, He, Li, Wang, and
  Yu}]{Peng2021HierarchicalTA}
Hao Peng, Jianxin Li, Qiran Gong, Senzhang Wang, Lifang He, Bo~Li, Lihong Wang,
  and Philip~S. Yu. 2021.
\newblock Hierarchical taxonomy-aware and attentional graph capsule rcnns for
  large-scale multi-label text classification.
\newblock \emph{IEEE Transactions on Knowledge and Data Engineering},
  33:2505--2519.

\bibitem[{Peng et~al.(2018)Peng, Li, He, Liu, Bao, Wang, Song, and
  Yang}]{Peng2018LargeScaleHT}
Hao Peng, Jianxin Li, Yu~He, Yaopeng Liu, Mengjiao Bao, Lihong Wang, Yangqiu
  Song, and Qiang Yang. 2018.
\newblock Large-scale hierarchical text classification with recursively
  regularized deep graph-cnn.
\newblock \emph{Proceedings of the 2018 World Wide Web Conference}.

\bibitem[{Pennington et~al.(2014)Pennington, Socher, and Manning}]{glove}
Jeffrey Pennington, Richard Socher, and Christopher~D. Manning. 2014.
\newblock Glove: Global vectors for word representation.
\newblock In \emph{Conference on Empirical Methods in Natural Language
  Processing}.

\bibitem[{Rojas et~al.(2020)Rojas, Bustamante, Oncevay, and
  Cabezudo}]{Rojas2020}
Kervy~Rivas Rojas, Gina Bustamante, Arturo Oncevay, and Marco
  Antonio~Sobrevilla Cabezudo. 2020.
\newblock \href {https://doi.org/10.18653/v1/2020.acl-main.205} {Efficient
  strategies for hierarchical text classification: External knowledge and
  auxiliary tasks}.
\newblock In \emph{Proceedings of the 58th Annual Meeting of the Association
  for Computational Linguistics, {ACL} 2020, Online, July 5-10, 2020}, pages
  2252--2257. Association for Computational Linguistics.

\bibitem[{{Sandhaus, Evan}(2008)}]{NYT}
{Sandhaus, Evan}. 2008.
\newblock \href {https://doi.org/10.35111/77BA-9X74} {The new york times
  annotated corpus}.

\bibitem[{Shannon(1948)}]{Shannon}
Claude~E. Shannon. 1948.
\newblock \href {https://doi.org/10.1002/j.1538-7305.1948.tb01338.x} {A
  mathematical theory of communication}.
\newblock \emph{Bell Syst. Tech. J.}, 27(3):379--423.

\bibitem[{Shimura et~al.(2018)Shimura, Li, and Fukumoto}]{Shimura2018HFTCNNLH}
Kazuya Shimura, Jiyi Li, and Fumiyo Fukumoto. 2018.
\newblock Hft-cnn: Learning hierarchical category structure for multi-label
  short text categorization.
\newblock In \emph{EMNLP}.

\bibitem[{Wang et~al.(2022{\natexlab{a}})Wang, Wang, Huang, Sun, and
  Wang}]{HGCLR}
Zihan Wang, Peiyi Wang, Lianzhe Huang, Xin Sun, and Houfeng Wang.
  2022{\natexlab{a}}.
\newblock \href {https://aclanthology.org/2022.acl-long.491} {Incorporating
  hierarchy into text encoder: a contrastive learning approach for hierarchical
  text classification}.
\newblock In \emph{Proceedings of the 60th Annual Meeting of the Association
  for Computational Linguistics (Volume 1: Long Papers), {ACL} 2022, Dublin,
  Ireland, May 22-27, 2022}, pages 7109--7119. Association for Computational
  Linguistics.

\bibitem[{Wang et~al.(2022{\natexlab{b}})Wang, Wang, Liu, Cao, Sui, and
  Wang}]{HPT}
Zihan Wang, Peiyi Wang, Tianyu Liu, Yunbo Cao, Zhifang Sui, and Houfeng Wang.
  2022{\natexlab{b}}.
\newblock Hpt: Hierarchy-aware prompt tuning for hierarchical text
  classification.

\bibitem[{Wu et~al.(2019)Wu, Xiong, and Wang}]{Wu2019Meta}
Jiawei Wu, Wenhan Xiong, and William~Yang Wang. 2019.
\newblock Learning to learn and predict: A meta-learning approach for
  multi-label classification.
\newblock In \emph{EMNLP}.

\bibitem[{Wu et~al.(2023)Wu, Chen, Shi, Li, and Xu}]{wu2023sega}
Junran Wu, Xueyuan Chen, Bowen Shi, Shangzhe Li, and Ke~Xu. 2023.
\newblock Sega: Structural entropy guided anchor view for graph contrastive
  learning.
\newblock In \emph{International Conference on Machine Learning}. PMLR.

\bibitem[{Wu et~al.(2022{\natexlab{a}})Wu, Chen, Xu, and Li}]{wu2022structural}
Junran Wu, Xueyuan Chen, Ke~Xu, and Shangzhe Li. 2022{\natexlab{a}}.
\newblock Structural entropy guided graph hierarchical pooling.
\newblock In \emph{International Conference on Machine Learning}, pages
  24017--24030. PMLR.

\bibitem[{Wu et~al.(2022{\natexlab{b}})Wu, Li, Li, Pan, and Xu}]{Wu2022HRN}
Junran Wu, Shangzhe Li, Jianhao Li, Yicheng Pan, and Keyulu Xu.
  2022{\natexlab{b}}.
\newblock A simple yet effective method for graph classification.
\newblock In \emph{IJCAI}.

\bibitem[{Xu et~al.(2019)Xu, Hu, Leskovec, and Jegelka}]{GIN}
Keyulu Xu, Weihua Hu, Jure Leskovec, and Stefanie Jegelka. 2019.
\newblock \href {https://openreview.net/forum?id=ryGs6iA5Km} {How powerful are
  graph neural networks?}
\newblock In \emph{7th International Conference on Learning Representations,
  {ICLR} 2019, New Orleans, LA, USA, May 6-9, 2019}. OpenReview.net.

\bibitem[{Yang et~al.(2018)Yang, Sun, Li, Ma, Wu, and Wang}]{Yang2018SGMSG}
Pengcheng Yang, Xu~Sun, Wei Li, Shuming Ma, Wei Wu, and Houfeng Wang. 2018.
\newblock Sgm: Sequence generation model for multi-label classification.
\newblock In \emph{COLING}.

\bibitem[{You et~al.(2019)You, Zhang, Wang, Dai, Mamitsuka, and
  Zhu}]{You2019AttentionXML}
Ronghui You, Zihan Zhang, Ziye Wang, Suyang Dai, Hiroshi Mamitsuka, and
  Shanfeng Zhu. 2019.
\newblock Attentionxml: Label tree-based attention-aware deep model for
  high-performance extreme multi-label text classification.
\newblock In \emph{NeurIPS}.

\bibitem[{Zhang et~al.(2022)Zhang, Zhu, Peng, Wu, and Xu}]{Zhang2022HINT}
Chong Zhang, He~Zhu, Xing~Qiang Peng, Junran Wu, and Ke~Xu. 2022.
\newblock Hierarchical information matters: Text classification via tree based
  graph neural network.
\newblock In \emph{COLING}.

\bibitem[{Zhou et~al.(2020)Zhou, Ma, Long, Xu, Ding, Zhang, Xie, and
  Liu}]{HiAGM}
Jie Zhou, Chunping Ma, Dingkun Long, Guangwei Xu, Ning Ding, Haoyu Zhang,
  Pengjun Xie, and Gongshen Liu. 2020.
\newblock Hierarchy-aware global model for hierarchical text classification.
\newblock In \emph{ACL}.

\end{thebibliography}
\bibliographystyle{acl_natbib}

\appendix
\section{Analysis of CIRCA}
\label{sec:appendix}
In this section, we first present the definition of coding tree following \citep{Li2016StructuralEntropy}. Secondly, we present the detailed flow of CIRCA, in particular, how each stage in Algorithm~\ref{alg:CIRCA} works, and the purpose of designing these steps. Finally, we give an analysis of the temporal complexity of CIRCA.

\paragraph{Coding Tree.} A coding tree $T$ of graph $G = {(V_G, E_G)}$ is defined as a tree with the following properties:
\begin{enumerate}[i.]
	\item For any node $v \in T$. $v$ is associated with a non-empty subset $V$ of $V_G$. Denote that $T_v = V$, in which $v$ is called the codeword of $V$ while $V$ (or $T_v$) is termed as the marker of $v$.\footnote{For simplicity, we do not distinguish the concept of node, codeword, and marker in the body of this paper.}
	 
	\item The coding tree has a unique root node $v_r$ that stands for the vertices set $V_G$ of $G$. That is, $T_{v_r} = V_G$.
	
	\item For every node $v \in T$, if $v_1, v_2, \dots, v_z$ are all the children of $v$, $\{T_{v_1}, T_{v_2}, \dots, T_{v_z}\}$ is a partition of $T_v$. That is, $T_v = \cup_{i=1}^{z} T_{v_i}$.
	
	\item For each leaf node $v_{\gamma} \in T$, $T_{v_{\gamma}}$ is a singleton. i.e. $v_{\gamma}$ corresponds to a unique node in $V_G$, and for any vertex $v \in V_G$, there is only one leaf node $v_{\tau} \in T$ that satisfies $T_{v_\tau} = v$.
\end{enumerate}

\paragraph{The workflow of CIRCA.}
In the initial state, the original graph $G = (V_G, E_G)$  is fed into CIRCA and each node in $V_G$ is treated as the leaf node of coding tree $T_L$ and directly linked with the root node $v_r$. The height of the initial coding tree is 1, which reflects the one-dimensional structure entropy of graph $G$. In other words, there are only two kinds of partition for $V_G$, one is the graph-level partition ($T_{v_r} = V_G$), and the other is the node-level partition ($T_{v_\tau} = v$).  We tend to find multi-granularity partitions for $G$, which could be provided by the $K$-dimensional optimal coding tree as the coding tree with height $K$ encodes and decodes $K + 1$ partitions in different levels for graph $G$.

In Stage~\ref{stg:1}, we merge the leaf nodes of the initial coding tree pair by pair until the root node $v_r$ has only two children. Merging leaf nodes is essentially compressing structural information, which is a process of reducing the structural entropy of graph $G$. When selecting the node pairs to be merged, we give priority to the nodes that reduce more structural entropy of graph $G$ after merging. 

After Stage~\ref{stg:1}, the coding tree $T$ becomes a binary tree, whose height is much greater than $K$ and closer to $log|V_G|$ in practical applications. In Stage~\ref{stg:2}, we tend to compress the coding tree $T$ to height $K$ by erasing its intermediate nodes. Note that removing nodes from the highly compressed coding tree is increasing the structural entropy of graph $G$. Thus, we preferentially erase the nodes that cause the minimal structural entropy increase.

The result of Stage~\ref{stg:2} might be an unbalanced tree that does not conform to the definition of coding trees. In Stage~\ref{stg:3}, we do some post-processing on the coding tree to make the leaf nodes the same height.


\paragraph{Complexity analysis.}
The time complexity of CIRCA is $O(h_{max}(|E_G|log|V_G| + |V_G|))$, where $h_{max}$ is the maximum height of coding tree $T_L$ during Stage~\ref{stg:1}. Since CIRCA tends to construct balanced coding trees, $h_{max}$ is no greater than $log(|V_G|)$.
\end{document}